\documentclass[sigconf,natbib]{acmart}
\AtBeginDocument{%
  }

\setcopyright{acmlicensed}
\copyrightyear{2018}
\acmYear{2018}
\acmDOI{XXXXXXX.XXXXXXX}
\acmConference[Conference acronym 'XX]{Make sure to enter the correct
  conference title from your rights confirmation email}{June 03--05,
  2018}{Woodstock, NY}
\acmISBN{978-1-4503-XXXX-X/2018/06}

\usepackage{algorithm,algorithmicx,algpseudocode}
\usepackage{multirow,booktabs}

\begin{document}

%%
%% The "title" command has an optional parameter,
%% allowing the author to define a "short title" to be used in page headers.
\title{STaTS: Structure-Aware Temporal Sequence Summarization via Statistical Window Merging}

%%
%% The "author" command and its associated commands are used to define
%% the authors and their affiliations.
%% Of note is the shared affiliation of the first two authors, and the
%% "authornote" and "authornotemark" commands
%% used to denote shared contribution to the research.
\author{Disharee Bhowmick}
\affiliation{%
  \institution{CSSE Department, Auburn University}
  \city{Auburn, AL}
  \country{USA}}
\email{san0028@auburn.edu}

\author{Ranjith Ramanathan}
\affiliation{%
  \institution{Department of Animal and Food Sciences, Oklahoma State University}
  \city{Stillwater, OK}
  \country{USA}}
\email{ranjith.ramanathan@okstate.edu}

\author{Sathyanarayanan N. Aakur}
\affiliation{%
  \institution{CSSE Department, Auburn University}
  \city{Auburn, AL}
  \country{USA}}
\email{san0028@auburn.edu}
% \author{Anonymous Submission}
% \affiliation{%
%   \institution{Anonymous}
%   \city{Anonymous}
%   \country{Anonymous}}
% \email{anonymous@submission.net}

%%
%% By default, the full list of authors will be used in the page
%% headers. Often, this list is too long, and will overlap
%% other information printed in the page headers. This command allows
%% the author to define a more concise list
%% of authors' names for this purpose.
\renewcommand{\shortauthors}{Bhowmick et al.}

%%
%% The abstract is a short summary of the work to be presented in the
%% article.
\begin{abstract}
Time series data often contain latent temporal structure—transitions between locally stationary regimes, repeated motifs, and bursts of variability—that are rarely leveraged in standard representation learning pipelines. Existing models typically operate on raw or fixed-window sequences, treating all time steps as equally informative, which leads to inefficiencies, poor robustness, and limited scalability in long or noisy sequences. We propose STaTS, a lightweight, unsupervised framework for Structure-Aware Temporal Summarization that adaptively compresses both univariate and multivariate time series into compact, information-preserving token sequences. STaTS detects change points across multiple temporal resolutions using a BIC-based statistical divergence criterion, then summarizes each segment using simple functions like the mean or generative models such as GMMs. This process achieves up to 30× sequence compression while retaining core temporal dynamics. STaTS operates as a model-agnostic preprocessor and can be integrated with existing unsupervised time series encoders without retraining. Extensive experiments on 150+ datasets—including classification tasks on the UCR-85, UCR-128, and UEA-30 archives, and forecasting on ETTh1/2, ETTm1, and Electricity—demonstrate that STaTS enables 85–90\% of the full-model performance while offering dramatic reductions in computational cost. Moreover, STaTS improves robustness under noise and preserves discriminative structure, outperforming uniform and clustering-based compression baselines. These results position STaTS as a principled, general-purpose solution for efficient, structure-aware time series modeling.

\end{abstract}

%%
%% The code below is generated by the tool at http://dl.acm.org/ccs.cfm.
%% Please copy and paste the code instead of the example below.
%%


%%
%% Keywords. The author(s) should pick words that accurately describe
%% the work being presented. Separate the keywords with commas.
%% A "teaser" image appears between the author and affiliation
%% information and the body of the document, and typically spans the
%% page.
\begin{teaserfigure}
  \includegraphics[width=\textwidth]{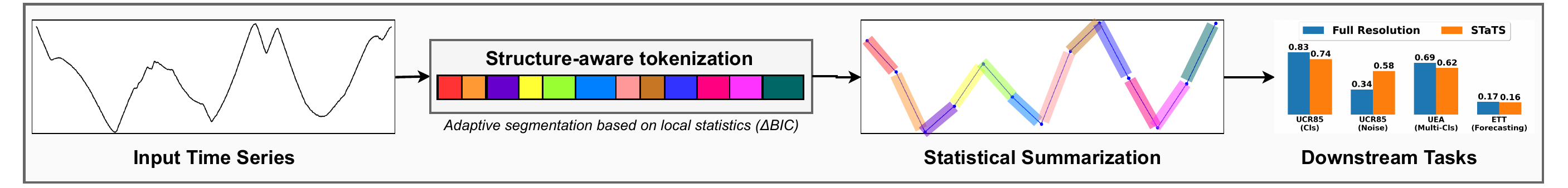}
  \caption{Overview of the STaTS Pipeline. Using a multi-scale BIC-based criterion, STaTS identifies statistically meaningful transitions using a raw input time series. It then performs structure-aware tokenization and summarizes each segment (e.g., by mean) to form a compact representation. These summaries enable efficient and robust downstream modeling, preserving task performance despite significant compression.
  % STaTS retains task performance across tasks despite substantial compression.
  }
  \Description{Overall STaTS Pipeline.}
  \label{fig:teaser}
\end{teaserfigure}

\received{20 February 2007}
\received[revised]{12 March 2009}
\received[accepted]{5 June 2009}

%%
%% This command processes the author and affiliation and title
%% information and builds the first part of the formatted document.
\maketitle

\section{Introduction}
Time series data is prevalent across diverse domains such as finance, Internet-of-Things (IoT), and healthcare, and continues to grow with advances in sensing technologies. As data collection capabilities expand, the length and complexity of recorded time series are increasing rapidly, placing significant computational demands on machine learning-based sequence understanding frameworks. These models typically operate on full-resolution inputs or apply fixed-size windows, treating all time steps equally informative. However, sequences often have a latent structure that captures essential details about the signal, such as transitions between locally stationary regimes, repeated motifs, and bursts of variability, that existing models often overlook. This assumption leads to inefficient processing and poor generalization, especially under noise, redundancy, or limited computing resources. These challenges raise a fundamental question: \textit{can time series be summarized to preserve task-relevant structure while enabling efficient and robust learning?}

Current approaches to time series representation learning typically fall into two extremes. Classical methods like Piecewise Aggregate Approximation (PAA)~\cite{keogh2000scaling}, SAX~\cite{lin2003symbolic}, and Dynamic Time Warping (DTW)~\cite{chen2013dtw} achieve efficient summarization but rely on uniform windowing or rigid symbolic encodings that ignore dynamic changes in signal complexity. Conversely, deep models such as TS2Vec~\cite{yue2022ts2vec} and TS-TCC~\cite{eldele2021time} process either complete sequences or apply sliding windows without regard for semantic variation. This lack of structure awareness leads to redundancy, computational overhead, and misalignment between the model’s tokenization and the signal’s true transitions, ultimately degrading representation quality. Additionally, fixed-window strategies risk over-segmenting stable regions and under-segmenting complex ones, compounding both inefficiency and error propagation. These limitations are particularly magnified under noisy conditions, where uniformly processed inputs tend to amplify spurious patterns and reduce generalization. Together, these challenges underscore the need for an adaptive summarization mechanism that aligns with the temporal heterogeneity of real-world signals and offers resilience to noise and variability.

To address these challenges, we propose \textit{STaTS} (Structure-Aware Tokenization for Time Series), a lightweight framework that adaptively compresses time series data by aligning summarization with the signal’s underlying structure. 
Figure~\ref{fig:teaser} illustrates the proposed pipeline. 
Rather than processing full-resolution sequences or applying uniformly spaced windows, STaTS first detects statistically significant change points using a BIC-inspired criterion across multiple temporal scales. These change points define dynamic segment boundaries that are each summarized—typically via the segment mean—into a compact representation. This design ensures that each token reflects a coherent region of the input sequence, allowing models to focus on semantically meaningful transitions rather than arbitrary intervals. The result is a drastically shorter input sequence that preserves salient dynamics while discarding redundancy and noise. Unlike learned attention or pooling modules, STaTS operates as a simple, unsupervised pre-processing step and can be paired with any downstream encoder. STaTS provides an efficient, interpretable, and robust alternative to conventional tokenization strategies by bridging statistical segmentation with modern representation learning.

\noindent Our \textbf{key contributions} are as follows:
\begin{itemize}
\item We propose \textbf{STaTS}, a structure-aware tokenization framework that identifies statistically coherent segments using a BIC-based change detection criterion applied across multiple temporal scales.
\item We introduce a modular and lightweight \textbf{summarization pipeline} that compresses time series by over 30× while preserving salient patterns, enabling efficient downstream modeling.
\item We show that STaTS operates in a \textbf{model-agnostic and unsupervised} fashion, requiring no architectural changes or gradient-based tuning, making it readily compatible with existing time series encoders such as TS2Vec.
\item We provide a unified interface for adapting STaTS to \textbf{classification, forecasting, and robustness tasks}, making it a general-purpose pre-processing tool for time series summarization.
\end{itemize}

We validate STaTS across three core tasks, univariate classification, multivariate classification, and long-horizon forecasting, on over 150 datasets from the UCR~\cite{bagnall_great_2017,chen2015ucr}, UEA~\cite{bagnall2018uea}, and ETT/Electricity benchmarks~\cite{zhou2021informer,dua2017uci}. Despite significant compression, integrating STaTS helps retain competitive performance with their full-resolution counterparts and outperforms uniform and clustering-based summarization baselines. In noisy settings, STaTS improves model robustness by filtering out spurious fluctuations, and in forecasting tasks, it preserves temporal dynamics across short and long horizons. 
These results affirm that STaTS offers a scalable and generalizable solution for efficient and robust time series modeling.

The remainder of this paper is organized as follows. Section~\ref{sec:related_work} reviews related work on time series summarization and representation learning. Section~\ref{sec:method} presents the STaTS framework, detailing the statistical segmentation strategy and summarization mechanisms. Section~\ref{sec:setup} describes the experimental setup, including datasets, baselines, and evaluation protocols. Section~\ref{sec:results} provides a comprehensive analysis of results across classification and forecasting tasks, including robustness and qualitative studies. Finally, Section~\ref{sec:conclusion} concludes with a discussion of key findings and future directions.

\section{Related Work}\label{sec:related_work}
\subsection{Time Series Classification}
Time series classification (TSC) aims to assign labels to time series based on their temporal patterns and structure. Recent progress in TSC spans a wide range of approaches, including interpretable feature-based methods, deep learning models, and ensembles. 
Classical techniques like shapelets and the Shapelet Transform introduced discriminative subsequences for interpretable classification~\cite{ye_time_2009,hills_classification_2014}, while symbolic models like BOSS leveraged symbolic Fourier approximation for robust performance in noisy settings~\cite{schafer_boss_2015}. Ensemble-based methods offer strong performance by integrating diverse classifiers to achieve state-of-the-art accuracy across many datasets~\cite{bagnall_great_2017,lemaire_usage_2020}. Deep learning-based frameworks provide end-to-end models such as FCN and ResNet~\cite{wang_time_2016}, with InceptionTime pushing scalability and accuracy further by ensembling Inception-style CNNs~\cite{ismail_fawaz_inceptiontime_2020}. Large-scale empirical studies benchmark deep models across 97 datasets and show that deep residual networks can rival ensembles in accuracy~\cite{fawaz_deep_2019}. Interval-based methods like r-STSF add interpretability by identifying discriminative sub-series~\cite{cabello_fast_2021}. For a deeper view of common TSC approaches and applications across domains like human activity recognition and earth observation, we recommend the reader to visit recent surveys\cite{foumani_deep_2023}.

\subsection{Time Series Forecasting}
Time series forecasting involves predicting future values of a sequence based on past observations. Deep learning has revolutionized this task by enabling data-driven models to learn expressive representations without relying on hand-crafted features. 
Early breakthroughs like sequence-to-sequence LSTMs established the foundation for end-to-end learning frameworks~\cite{sutskever_sequence_2014}, while models such as DeepAR extended autoregressive forecasting to large collections of related time series with probabilistic outputs~\cite{salinas_deepar_2019}. N-BEATS introduced a residual MLP architecture for accurate and interpretable univariate forecasts~\cite{oreshkin_n-beats_2020}, and the Temporal Fusion Transformer (TFT) incorporated attention and gating mechanisms for state-of-the-art multi-horizon prediction~\cite{lim_temporal_2020}. More recent models like CARD capture both temporal and cross-channel dependencies using channel-aligned attention~\cite{wang_time_2016}, while generative models like WaveNet demonstrated powerful autoregressive modeling via dilated convolutions~\cite{oord_wavenet_2016}. The growing ecosystem of tools (e.g., GluonTS~\cite{alexandrov_gluonts_2019}) and surveys~\cite{faloutsos_forecasting_2018, kim_comprehensive_2025} highlight the field’s evolution toward increasingly scalable, interpretable, and robust forecasting systems, while also identifying challenges such as distribution shift, causality, and channel dependency. 

\subsection{Time series summarization.} 
Early approaches to time series summarization include Piecewise Aggregate Approximation (PAA)~\cite{keogh2000scaling}, which partitions univariate sequences into equal-length segments and replaces each with its mean. Symbolic Aggregate approXimation (SAX)~\cite{lin2003symbolic} builds on PAA by quantizing the segment means into discrete symbols, enabling symbolic indexing and similarity search. However, both methods assume fixed-length segmentation and operate only on univariate data, limiting their applicability to complex, high-dimensional settings. More recent methods, such as Toeplitz Inverse Covariance Clustering (TICC)~\cite{hallac2017toeplitz}, extend to multivariate segmentation by fitting sparse graphical models to fixed-length sliding windows, but require solving a computationally expensive optimization problem and are not designed for efficient input compression or downstream model integration. In contrast, STaTS introduces a lightweight, model-agnostic framework for summarizing multivariate time series by detecting statistically coherent segments across multiple temporal resolutions using a full-covariance BIC criterion. Each segment is then mapped to a compact token via a simple summarization function, yielding a variable-length, structure-preserving representation that any downstream time series model can directly consume.

% \clearpage
\section{Methodology}\label{sec:method}
We now present the STaTS framework for structure-aware temporal summarization. STaTS operates in two stages: first, it segments the input sequence into statistically coherent chunks using a multi-resolution change point detection strategy based on the Bayesian Information Criterion (BIC); then, it summarizes each segment into a compact representation using simple statistical functions or generative models. This two-step process yields a compressed sequence that preserves core temporal structure while reducing redundancy and noise. In this section, we describe the segmentation and summarization components in detail and outline how STaTS integrates seamlessly with downstream representation learning models.

\begin{figure}[t]
    \centering
    \includegraphics[width=0.95\linewidth]{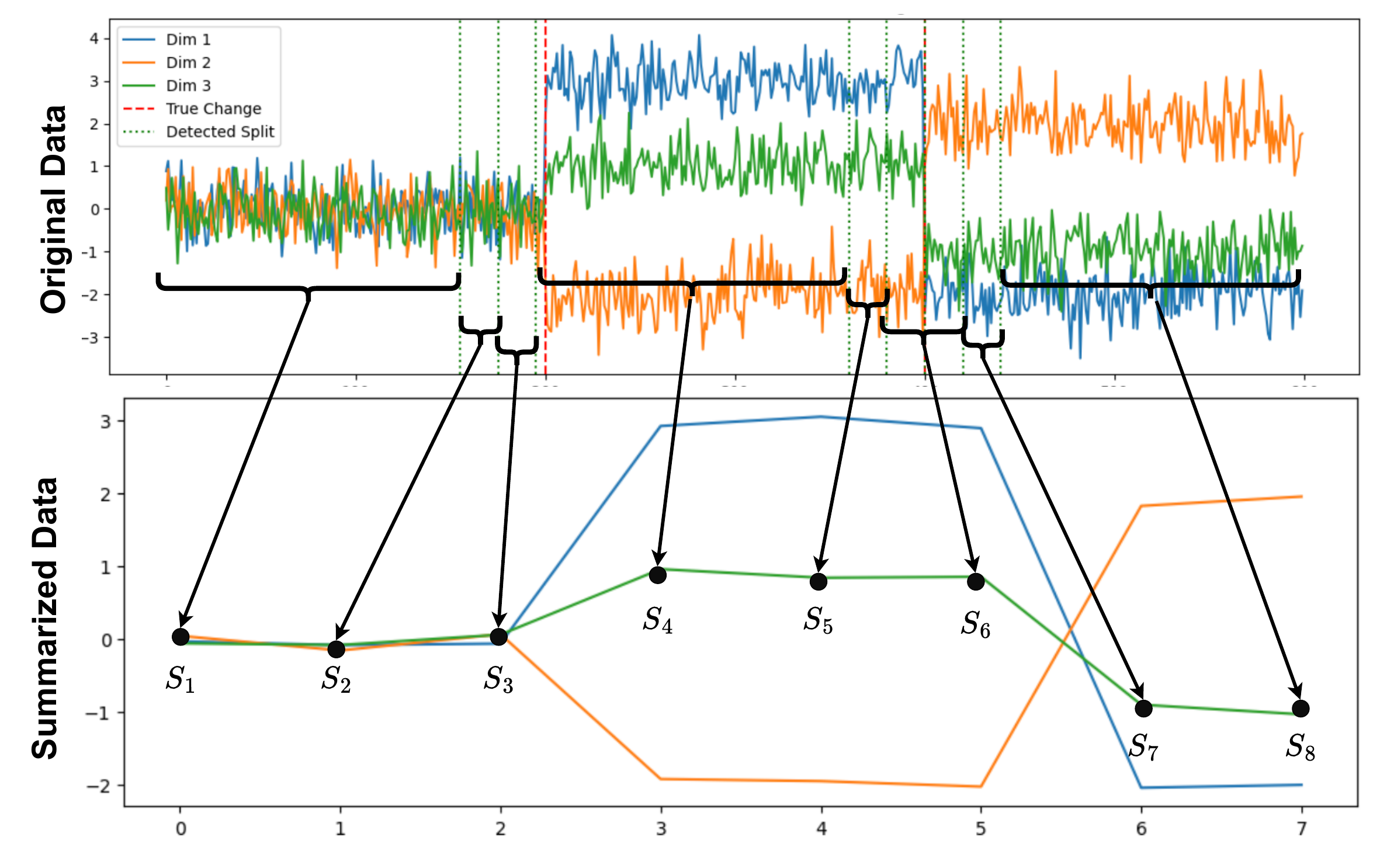}
    \caption{\textit{Visual overview of STaTS-based summarization.} (Top) A multivariate time series with detected splits and true change points. (Bottom) The summarized sequence, where each token $S_i$ represents a segment. STaTS does not aim to recover true changes; segments are detected and summarized to capture structure relevant for downstream tasks.}
    \label{fig:overview}
    \Description[STaTS summarization process]{A two-panel diagram showing the original multivariate time series (top) with detected splits and ground-truth change points, and the summarized sequence (bottom) with tokens representing each segment.}
\end{figure}

\subsection{Overview} 
Given a multivariate time series $\mathbf{X} \in \mathbb{R}^{T \times d}$, where $T$ is the number of timesteps and $d$ is the number of dimensions, our goal is to transform $\mathbf{X}$ into a much shorter sequence $\tilde{\mathbf{X}} \in \mathbb{R}^{T' \times d}$, where $T' \ll T$, such that the resulting sequence retains the underlying structure necessary for downstream tasks like classification or forecasting. We refer to this process as \textit{structure-aware temporal summarization}. 
We first segment $\mathbf{X}$ into a sequence of statistically coherent subsequences, or \textit{segments}, based on multivariate similarity. Each segment is then mapped to a representative \textit{summary token}, resulting in a condensed sequence that captures key statistical characteristics of the original data. 
Formally, we identify a set of split points $\{ \tau_1, \ldots, \tau_{T'-1} \}$ that divide the input into $T'$ non-overlapping segments:
$S_i = \mathbf{X}[\tau_{i-1}:\tau_i], \quad \text{where } \tau_0 = 0 \text{ and } \tau_{T'} = T.$
Each segment $S_i \in \mathbb{R}^{(\tau_i - \tau_{i-1}) \times d}$ is then summarized using a function $\phi(S_i) \in \mathbb{R}^d$, such as its mean pooling. 
The final summarized sequence is:
\begin{equation}
\tilde{\mathbf{X}} = [\phi(S_1), \ldots, \phi(S_{T'})] \in \mathbb{R}^{T' \times d}.
\label{eqn:tokens_defn}
\end{equation} 
We refer to each vector $\phi(S_i)$ as a \textit{summary token}, and to the overall result $\tilde{\mathbf{X}}$ as the \textit{tokenized summary}. This sequence can be used as a drop-in replacement for the original input in any downstream model, enabling significant reductions in sequence length and computational cost without sacrificing predictive performance. 
Figure~\ref{fig:overview} illustrates this process using a toy example. The top panel shows a multivariate time series with three dimensions, along with detected statistical splits (green) and ground-truth changes (red). Each detected segment is mapped to a token $S_i$, shown in the bottom panel. While some splits align with the true change points, the goal of STaTS is not to detect events, but to capture coherent statistical structure for effective summarization.

\begin{algorithm}[t]
\caption{STaTS: Multi-Scale Tokenization and Summarization for Time Series}
\label{alg:stats}
\textbf{Input}: Time series $\mathbf{X} \in \mathbb{R}^{T \times d}$, window sizes $\Delta = \{ \delta_{\min}, \ldots, \delta_{\max} \}$, threshold multiplier $\alpha$, minimum separation $s_{\min}$ \\
\textbf{Output}: Tokenized summary $\tilde{\mathbf{X}} \in \mathbb{R}^{T' \times d}$
\begin{algorithmic}[1]
\State \textbf{Initialize:} Candidate splits $\mathcal{C} \leftarrow \emptyset$
\For{each $\delta \in \Delta$} \Comment{\textit{Evaluate across multiple resolutions}}
    \For{each pair of adjacent windows $x_1, x_2 \in \mathbb{R}^{\delta \times d}$}
        \State Compute $\Delta \mathrm{BIC}$ using Equation~\ref{eqn:bic_split}
    \EndFor
    \State Compute mean $\mu_\delta$ and std. dev. $\sigma_\delta$ of scores
    \For{each position $t$ with score $s_t$}
        \If{ $s_t \geq \mu_\delta + \alpha \cdot \sigma_\delta$ }
            \State Add $t$ to $\mathcal{C}$
        \EndIf
    \EndFor
\EndFor
\State Apply non-maximum suppression to $\mathcal{C}$ with minimum separation $s_{\min}$
\State Define split points $\{ \tau_1, \ldots, \tau_{T'-1} \}$, with $\tau_0 = 0$, $\tau_{T'} = T$
\For{each segment $S_i = \mathbf{X}[\tau_{i-1}:\tau_i]$}
    \State Compute summary token $\phi(S_i) = \frac{1}{|S_i|} \sum_{t=\tau_{i-1}}^{\tau_i - 1} \mathbf{x}_t$
\EndFor
\State \textbf{Return:} $\tilde{\mathbf{X}} = [\phi(S_1), \ldots, \phi(S_{T'})]$
\end{algorithmic}
\end{algorithm}

% Algorithm~\ref{alg:stats} summarizes the end-to-end STaTS pipeline. Lines 1–10 correspond to the tokenization phase (detailed in Section~\ref{sec:tokenization}), where the input sequence is evaluated across a dynamic range of window sizes to identify statistically significant change points using the BIC-based criterion in Equation~\ref{eqn:bic_split}. Each window scale is thresholded independently, and candidate splits are refined through non-maximum suppression to ensure sparsity and robustness. Lines 11–13 represent the summarization phase (Section~\ref{sec:summarization}), where each detected segment is mapped to a fixed-length token using a summarization function \( \phi \), such as the segment mean. The resulting sequence of summary tokens \( \tilde{\mathbf{X}} \) provides a compact, structure-preserving representation of the original input, enabling efficient downstream modeling.

Algorithm~\ref{alg:stats} formalizes the STaTS pipeline as a two-stage process: (1) statistically adaptive tokenization and (2) summarization. In the tokenization phase (Lines 1–10), the input time series \( \mathbf{X} \in \mathbb{R}^{T \times d} \) is scanned at multiple window scales \( \delta \in \Delta \) to detect distributional shifts. For each candidate split point \( t \), we compute the change in Bayesian Information Criterion (BIC), defined as:
\[
\Delta \text{BIC}(t) = \text{BIC}(\mathbf{X}_{1:t}) + \text{BIC}(\mathbf{X}_{t+1:T}) - \text{BIC}(\mathbf{X}),
\]
where each segment's BIC is computed under the assumption of multivariate Gaussianity:
\[
\text{BIC}(\mathbf{X}) = \frac{n}{2} \log |\Sigma| + \lambda \cdot k \log n,\quad \text{with } k = d + \frac{d(d+1)}{2}.
\]
A point \( t \) is marked as a candidate boundary if \( \Delta \text{BIC}(t) \) exceeds the adaptive threshold \( \mu_\delta + \alpha \cdot \sigma_\delta \) at scale \( \delta \), where \( \mu_\delta \) and \( \sigma_\delta \) are the mean and standard deviation of the scores at that resolution. Final split points are obtained via non-maximum suppression using a minimum separation \( s_{\min} \). In the summarization phase (Lines 11–13), each segment is mapped to a fixed-length token via a summarization function \( \phi \), such as the mean, yielding a compressed representation \( \tilde{\mathbf{X}} \in \mathbb{R}^{T' \times d} \). This representation preserves structural transitions while dramatically reducing sequence length, enabling efficient downstream modeling.

\subsection{Tokenization}\label{sec:tokenization}

The tokenization process in STaTS is based on identifying statistically coherent regions of a time series by evaluating the likelihood of distributional similarity across adjacent temporal windows. 
For a pair of adjacent windows $x_1, x_2 \in \mathbb{R}^{\delta \times d}$, where $\delta$ is the window size and $d$ is the number of dimensions, we estimate their empirical covariance matrices and assess whether their combined distribution is better modeled as a single multivariate Gaussian or as two separate ones. 
This is formalized as a penalized likelihood comparison, where we contrast the negative log-likelihood of the joint model against the sum of the individual models. Specifically, we compute:

\begin{equation}
\Delta \mathrm{BIC} = -2 \left( \ell_{\mathrm{joint}} - \ell_{\mathrm{sep}} \right) + k \log(2\delta)    
\label{eqn:bic_split}
\end{equation}

where $\ell_{\mathrm{sep}} = -\frac{\delta}{2} \left( \log |\Sigma_1| + \log |\Sigma_2| \right)$, $\ell_{\mathrm{joint}} = -\delta \log |\Sigma_{12}|$, and $k = d + \frac{d(d+1)}{2}$ denotes the number of free parameters in the full covariance model. Here, $\Sigma_1$, $\Sigma_2$, and $\Sigma_{12}$ refer to the empirical covariances of $x_1$, $x_2$, and their concatenation, respectively.

While we evaluate $\Delta \mathrm{BIC}$ locally to detect candidate splits, the tokenization procedure can be interpreted as approximating a global penalized likelihood objective over the entire sequence. Given a set of segment boundaries $\{ \tau_0, \tau_1, \ldots, \tau_{T'} \}$ and corresponding segments $S_i = \mathbf{X}[\tau_{i-1}:\tau_i]$, the objective becomes minimizing the overall segmentation cost:

\begin{equation}
\mathcal{L}_{\mathrm{BIC}}(\{S_i\}) = \sum_{i=1}^{T'} \left( -\frac{|S_i|}{2} \log |\Sigma_i| + \frac{k}{2} \log |S_i| \right)
\label{eqn:bic_objective}
\end{equation}

where $\Sigma_i$ is the empirical covariance of segment $S_i$, $|S_i| = \tau_i - \tau_{i-1}$, and $k = d + \frac{d(d+1)}{2}$ as before. The first term encourages segments with high internal statistical consistency, while the second penalizes over-segmentation through a model complexity term. In this view, tokenization corresponds to finding a segmentation that minimizes $\mathcal{L}_{\mathrm{BIC}}$, striking a balance between statistical fit and model parsimony. 
Unlike prior work that applied BIC-based segmentation in univariate settings~\cite{vavilthota2024capturing}, our formulation generalizes naturally to multivariate time series through full covariance matrices. This enables STaTS to account for inter-dimensional dependencies when assessing statistical homogeneity, making it broadly applicable to high-dimensional time series data.

\paragraph{Multi-scale Coherence Detection.}
Rather than relying on a fixed window size, STaTS evaluates statistical coherence across a range of temporal resolutions. For each value of $\delta$ in a predefined range, we compute the BIC-based change score from Equation~\ref{eqn:bic_split} by comparing adjacent windows of length $\delta$. This multi-scale evaluation allows STaTS to detect changes that manifest at different timescales, capturing both short, abrupt, and longer, gradual shifts. For each \(\delta\), we identify candidate split points by thresholding the distribution of scores using $\mu_\delta + \alpha \cdot \sigma_\delta$, where $\mu_\delta$ and $\sigma_\delta$ are the mean and standard deviation of the \(\Delta \mathrm{BIC}\) scores at that scale. The resulting candidates are aggregated across all window sizes, and redundant detections are pruned via non-maximum suppression. This dynamic windowing strategy improves robustness to noise and enables flexible segment granularity while maintaining the theoretical foundation of the underlying statistical model.

\subsection{Summarization}\label{sec:summarization}

Once the time series has been segmented into statistically coherent regions $\{S_1, \ldots, S_{T'}\}$ through tokenization, the next step is to summarize each segment into a compact representation. The goal of summarization is to reduce redundancy within each segment while preserving its statistical characteristics for downstream modeling.

We define a summarization function $\phi: \mathbb{R}^{n \times d} \rightarrow \mathbb{R}^d$, which maps each segment $S_i \in \mathbb{R}^{|S_i| \times d}$ to a fixed-length token $\phi(S_i) \in \mathbb{R}^d$. In this work, we adopt mean pooling as the default summarization operator:
\begin{equation}
\phi(S_i) = \frac{1}{|S_i|} \sum_{t=\tau_{i-1}}^{\tau_i - 1} \mathbf{x}_t
\label{eqn:summarization}
\end{equation}
This operation captures the segment’s first-order statistical properties and introduces minimal distortion to downstream tasks. Since segments are constructed to be statistically homogeneous, the mean provides a natural and information-preserving summary. 
Furthermore, the mean vector corresponds to the maximum likelihood estimate (MLE) of the segment's central tendency under the full-covariance Gaussian model assumed in Equation~\ref{eqn:bic_split}, reinforcing its statistical validity as a representative token. 
Although we use mean pooling for its efficiency and simplicity, the summarization function $\phi$ is easily extensible. Alternative formulations may use principal components, cluster centroids, or samples from generative models to provide richer representations. This flexibility allows STaTS to adapt to a range of modeling requirements and domain-specific constraints. 
The output of this stage is a condensed sequence $\tilde{\mathbf{X}} = [\phi(S_1), \ldots, \phi(S_{T'})] \in \mathbb{R}^{T' \times d}$, which serves as the tokenized summary of the original time series and can be used directly in downstream models. 
Because segments are not uniformly long, the resulting tokens offer an adaptive summarization of the signal—short segments capture fast-changing regions, while longer segments compress more stable intervals. This leads to a temporal abstraction that better reflects the underlying dynamics of the data.

\subsubsection{Statistical Validity}
Under the assumption that each segment $S_i$ is generated from an independent multivariate Gaussian distribution, the segment mean $\phi(S_i)$ is a sufficient statistic for the underlying distribution. That is, the likelihood of the segment depends only on its empirical mean and covariance, both of which are captured or approximated in our summarization. Since tokenization ensures that each segment is statistically coherent, the use of the mean as a summary retains the key discriminative properties of the original data, particularly for models sensitive to first-order structure.

\subsubsection{Computational Efficiency.}
Let $C_{\text{model}}(T)$ denote the cost of running a downstream model on a sequence of length $T$. For sequence models like TS2Vec~\cite{yue2022ts2vec}, this cost typically scales linearly or quadratically with sequence length. By reducing the sequence length from $T$ to $T' \ll T$, STaTS reduces the number of floating-point operations (FLOPs) by a factor of $T / T'$, yielding an $\mathcal{O}(T/T')$ speedup in inference and training. Empirically, we find that STaTS reduces input length by approximately 10×, resulting in comparable gains in computational efficiency without significant degradation in predictive performance.

\subsection{Implementation Details}

All input time series are first normalized to zero mean and unit variance. For tokenization, STaTS evaluates statistical coherence across a range of window sizes $\delta \in \{ \delta_{\min}, \delta_{\min} + \Delta\delta, \ldots, \delta_{\max} \}$, where we use $\delta_{\min} = 5$, $\delta_{\max} = 500$, and a step size of $\Delta\delta = 5$. Adjacent windows are compared with a stride of \texttt{10} to balance granularity and efficiency. For each window pair, full covariance matrices are computed and regularized for numerical stability by adding $\epsilon \cdot I$, where $\epsilon = 10^{-6}$. Change points are selected based on BIC scores exceeding an adaptive threshold $\mu_\delta + \alpha \cdot \sigma_\delta$, with \(\alpha\) set to 2. To prevent redundant detections, non-maximum suppression is applied with a minimum separation of $s_{\min} = 20$ timesteps. Each resulting segment is summarized using mean pooling by default, though STaTS supports other summarization functions such as GMM sampling. In all experiments, STaTS achieves an average sequence length reduction of approximately $33\times$ on univariate data and $20\times$ on multivariate data, leading to a corresponding reduction of $20\times$ in downstream model computation. The summarized sequences are model-agnostic and can be passed directly to standard time series models such as TS2Vec without requiring architectural changes or retraining.

\section{Experimental Setup}\label{sec:setup}
To evaluate the effectiveness and generality of STaTS, we conduct extensive experiments across major time series tasks: univariate and multivariate classification and forecasting. Our goal is to assess how well STaTS-based summarization preserves task-relevant information under substantial compression, and how it compares to full-resolution baselines and alternative summarization strategies. We describe the datasets used for each task, the baselines we compare against, and the evaluation metrics employed to quantify both accuracy and efficiency.

\begin{table}[t]
\centering
\resizebox{0.475\textwidth}{!}{  
\begin{tabular}{lcccccc}
\toprule
\multirow{2}{*}{Model} & \multicolumn{3}{c}{UCR-85} & \multicolumn{3}{c}{UCR-128} \\
\cmidrule(lr){2-4} \cmidrule(lr){5-7}
 & Accuracy & Rank & Avg. Length & Accuracy & Rank & Avg. Length \\
\midrule
T-Loss             & 0.805 & 2.98 & 424.4 & 0.806 & 3.02 & 534.5 \\
TNC                & 0.779 & 3.75 & 424.4 & 0.761 & 4.10 & 534.5 \\
TS-TCC             & 0.778 & 3.78 & 424.4 & 0.757 & 4.10 & 534.5 \\
TST                & 0.649 & 6.94 & 424.4 & 0.639 & 6.76 & 534.5 \\
DTW                & 0.740 & 5.18 & 424.4 & 0.728 & 5.29 & 534.5 \\
TS2Vec             & \textbf{0.829} & \textbf{1.99} & 424.4 & \textbf{0.829} & \textbf{2.02} & 534.5 \\
\midrule
TS2Vec (mean)      & 0.739 & 4.82 & 12.1  & 0.741 & 4.39 & 12.9 \\
TS2Vec (GMM)       & 0.655 & 7.35 & 60.7  & 0.664 & 6.92 & 73.2 \\
TS2Vec (uniform)   & 0.621 & 8.21 & 12.1  & 0.616 & 8.10 & 12.9 \\
\bottomrule
\end{tabular}
}
\caption{
Performance comparison across the UCR-85 and UCR-128 archives. The last block includes TS2Vec variants that operate on compressed representations. 
\textbf{TS2Vec (mean)} applies STaTS-based adaptive summarization, achieving strong performance while reducing input length by over 33$\times$. 
Alternative summarization strategies, such as uniform segmentation and GMM-based sampling, perform significantly worse in both accuracy and average rank.
}
\label{tab:model_compression_comparison}
\end{table}
\subsection{Tasks and Data}
We evaluate the impact of statistical summarization across three core time series learning tasks: univariate classification, multivariate classification, and multivariate forecasting. For univariate classification, we use the UCR-128~\cite{dau2019ucr} and UCR-85~\cite{chen2015ucr} archives, which span a wide range of domains such as sensor signals, ECG, spectrographs, and handwritten shapes. The UCR-128 collection contains 128 diverse datasets with fixed-length sequences and class labels, while UCR-85 is a curated subset with broader adoption in recent representation learning benchmarks. For multivariate classification, we adopt the UEA-30~\cite{bagnall2018uea} archive, which includes higher-dimensional datasets from wearable sensors, motion capture, and medical devices, often with varying numbers of input channels and richer temporal dynamics. To assess the generality of our approach in predictive modeling, we also evaluate on multivariate forecasting, using four benchmark datasets: ETTh1~\cite{zhou2021informer}, ETTh2~\cite{zhou2021informer}, ETTm1~\cite{zhou2021informer}, and Electricity~\cite{dua2017uci}, following prior work~\cite{yue2022ts2vec}. These datasets cover both hourly and minute-level resolutions and present forecasting challenges at varying horizons, ranging from short-term (24 steps) to long-term (720 steps). Together, these benchmarks allow us to rigorously assess the ability of compressed representations to support accurate, robust, and generalizable time series modeling.

\subsubsection{Metrics}
We evaluate each task using metrics reflecting performance and comparability across datasets. For classification (both univariate and multivariate), we report the average accuracy across datasets and the average rank of each method, where lower ranks indicate better performance. Accuracy directly measures correct predictions, while rank accounts for relative performance across datasets of varying difficulty and scale. For forecasting, we use normalized Mean Squared Error (nMSE), which divides the raw MSE by the variance of the target series, allowing fair comparison across datasets with different magnitudes. These metrics are consistent with prior work~\cite{yue2022ts2vec} and capture absolute error and method robustness across diverse settings.

\subsection{Baselines}
Our primary focus is on evaluating the effect of statistical summarization within the TS2Vec framework, which serves as the backbone for all our experiments. 
% TS2Vec is a contrastive learning-based model that learns universal time series representations by maximizing temporal alignment across multiple resolutions using dilated 1D convolutional encoders. 
We compare our proposed summarization approach, TS2Vec (mean), against several variants and existing baselines. 
TS2Vec (GMM) replaces mean-based summarization with a Gaussian Mixture Model. It fits a 5-component GMM to each segment and concatenates the parameters (means and variances) to form a fixed-size representation. 
In contrast, TS2Vec (uniform) skips statistical segmentation altogether, dividing the input into 10 equally sized chunks, a number that approximates the average number of segments discovered by STaTS across datasets. 
Thus, TS2Vec (uniform) can be viewed as TS2Vec applied on PAA-compressed inputs, where the original time series is divided into 10 equal-length segments and each segment is represented by its mean, similar in spirit to classical Piecewise Aggregate Approximation (PAA)~\cite{keogh2000scaling}. 
For classification, we also compare against established time series classification baselines: T-Loss~\cite{franceschi2019unsupervised}, which uses triplet loss on random positive and negative subseries; TNC (Temporal Neighborhood Coding)~\cite{tonekaboni2021unsupervised}, which predicts whether two subseries belong to the same temporal neighborhood; and TS-TCC~\cite{eldele2021time}, which learns temporal alignment via positive-negative pair classification. We include TST (Time Series Transformer)~\cite{zerveas2021transformer} as a transformer-based architecture and DTW (Dynamic Time Warping)~\cite{chen2013dtw} as a non-learned similarity-based baseline. For forecasting, we benchmark against Informer~\cite{zhou2021informer}, which uses a sparse self-attention mechanism for efficient long-range forecasting, and TCN (Temporal Convolutional Network)~\cite{bai2018empirical}, which uses dilated convolutions to model temporal dependencies. 
These baselines allow us to assess how well compressed representations retain task-relevant information compared to full-resolution and transformer-style models.

\section{Results and Discussion}\label{sec:results}
This section presents results demonstrating the effectiveness of STaTS across the tasks introduced earlier. Our evaluation focuses on three key dimensions: (1) how well STaTS-based summarization preserves downstream performance under compression, (2) its robustness in noisy or long-horizon settings, and (3) how it compares to alternative compression strategies and full-resolution models. We highlight both quantitative gains and qualitative patterns to show that STaTS offers a practical trade-off between efficiency and accuracy, enabling scalable time series learning.

\begin{table}[t]
\centering
\resizebox{0.475\textwidth}{!}{  
\begin{tabular}{lcccccc}
\toprule
\multirow{2}{*}{Model} & \multicolumn{3}{c}{UCR-85 (noisy)} & \multicolumn{3}{c}{UCR-128 (noisy)} \\
\cmidrule(lr){2-4} \cmidrule(lr){5-7}
 & Accuracy & Rank & Avg. Length & Accuracy & Rank & Avg. Length \\
\midrule
TS2Vec (ori)       & 0.336 & 3.24 & 424.4 & 0.412 & 2.88 & 534.5 \\
TS2Vec (uniform)   & 0.475 & 3.00 & 12.1  & 0.485 & 3.15 & 12.9 \\
TS2Vec (GMM)       & 0.505 & 2.53 & 60.7  & 0.522 & 2.66 & 73.2 \\
TS2Vec (mean)      & \textbf{0.581} & \textbf{1.23} & 12.1  & \textbf{0.603} & \textbf{1.32} & 12.9 \\
\bottomrule
\end{tabular}
}
\caption{
Robustness of TS2Vec variants under additive Gaussian noise across UCR-85 and UCR-128. 
\textbf{TS2Vec (mean)}, using STaTS-based adaptive summarization, achieves the highest accuracy and lowest rank in both settings, 
outperforming GMM- and uniformly-segmented models, as well as the full-resolution baseline.
}
\label{tab:robustness_noise}
\end{table}

\subsection{Univariate Classification}

First, we evaluate the performance of STaTS on the univariate classification task. Table~\ref{tab:model_compression_comparison} summarizes the results, where we compare against strong unsupervised time series classification baselines. 

\textbf{UCR-128 Archive.} On the full UCR-128 archive, TS2Vec (ori) achieves the highest classification accuracy (0.829) and the best average rank (2.02), serving as a strong backbone model for uncompressed inputs. Among compressed variants, TS2Vec (mean) retains a strong performance of 0.741 accuracy and a rank of 4.39, despite operating on inputs that are over 33× shorter. This translates to retaining nearly 90\% of the original performance while drastically reducing the computational footprint. In contrast, TS2Vec (uniform) and TS2Vec (GMM) suffer sharper degradation, with accuracies of 0.616 and 0.664 respectively, and notably higher ranks (8.10 and 6.92). When analyzing dataset distribution, UCR-128 contains a substantial number of very long sequences ($>$700) (35 out of 90 datasets), where uniform segmentation becomes increasingly brittle and prone to misalignment with semantically meaningful events. GMM-based summarization also underperforms here, likely due to unstable or ill-posed clustering on high-dimensional or low-variance segments. STaTS, by contrast, dynamically adapts chunk boundaries based on local statistical regularities, enabling it to compress without flattening or distorting key transitions in the signal. These findings suggest that STaTS provides not only an efficient compression scheme, but also a robust inductive bias for long-sequence understanding, with consistent performance across a wide range of sequence lengths and domains.

\textbf{UCR-85 Archive.} In the UCR-85 benchmark, where datasets are more evenly distributed across sequence lengths—with 19 short ($\leq$100), 23 medium (101--300), 25 long (301--700), and 17 very long ($>$700) datasets—TS2Vec (ori) again leads with the highest accuracy (0.829) and lowest rank (1.99). Here, TS2Vec (mean) remains competitive with 0.739 accuracy and a rank of 4.82, outperforming all compressed variants. The performance gap between STaTS-based summarization and alternative compression strategies is even more pronounced in this setting: TS2Vec (GMM) and TS2Vec (uniform) achieve only 0.655 and 0.621 accuracy, with ranks of 7.35 and 8.21 respectively. These results confirm that STaTS is particularly effective for short and medium-length sequences, which together account for over 50\% of UCR-85. On such sequences, STaTS avoids over-segmentation by adapting to local variance, while uniform splitting introduces artificial boundaries and GMM fitting can be unstable due to limited segment length. Importantly, STaTS generalizes well across the entire length spectrum—unlike uniform segmentation, which assumes equidistant structure, or GMM, which struggles with underdetermined clusters. The nearly 3.5-point rank advantage over uniform segmentation underscores STaTS’s ability to identify task-relevant patterns, even in highly compressed form. These findings reinforce the notion that summarization must be structure-aware, not just length-aware, to preserve fidelity in fine-grained classification tasks.

\begin{table}[t]
\centering
% \resizebox{0.475\textwidth}{!}{  
\begin{tabular}{lccc}
\toprule
Model & Accuracy & Rank & Avg. Length \\
\midrule
TS2Vec (ori)       & \textbf{0.690} & \textbf{3.39} & 163.4 \\
TNC                & 0.665          & 4.78          & 163.4 \\
TS-TCC             & 0.653          & 4.67          & 163.4 \\
T-Loss             & 0.644          & 4.11          & 163.4 \\
DTW                & 0.641          & 4.90          & 163.4 \\
TST                & 0.607          & 5.52          & 163.4 \\
\midrule
TS2Vec (mean)      & 0.622          & 4.70          & 8.2 \\
TS2Vec (GMM)       & 0.589          & 5.52          & 16.3 \\
TS2Vec (uniform)   & 0.488          & 7.26          & 8.2 \\
\bottomrule
\end{tabular}
% }
\caption{
Performance comparison on the UEA-30 multivariate dataset. 
\textbf{TS2Vec (mean)} with STaTS-based adaptive summarization outperforms both GMM- and uniformly segmented versions of TS2Vec by large margins. All compressed variants operate on sequences ~20$\times$ shorter.
}
\label{tab:uea_compression_comparison}
\end{table}

\textbf{Robustness Under Noise.} 
The results in Table~\ref{tab:robustness_noise} demonstrate a striking shift in performance dynamics when additive Gaussian noise is introduced to the inputs. Under these conditions, TS2Vec (mean)—despite operating on sequences compressed by over 30×—not only maintains its competitive edge but outperforms the full-resolution TS2Vec model by a substantial margin. Specifically, it achieves the highest accuracy and lowest rank across both UCR-85 (0.581 accuracy, 1.23 rank) and UCR-128 (0.603 accuracy, 1.32 rank), indicating that STaTS-based summarization confers significant robustness to input corruption. In contrast, the original TS2Vec model suffers the most under noise, dropping to 0.336 and 0.412 accuracy on UCR-85 and UCR-128 respectively, a relative degradation of nearly 50\%. TS2Vec (uniform) also shows notable sensitivity, degrading by over 24\% on UCR-85 compared to its clean setting. 
This robust performance can be attributed to the implicit denoising effect of STaTS. The summarization process smooths over noise by aggregating local statistics within coherent segments while preserving salient structural variation. Uniform segmentation and GMM-based summarization offer moderate resilience but lack the adaptive nature of STaTS, resulting in suboptimal chunk boundaries or unstable parameter fits under noise. These findings underscore a crucial benefit of statistical summarization: it not only enables compression but also acts as a powerful inductive prior for robustness, particularly in real-world scenarios where data may be noisy, incomplete, or corrupted. With the reduced FLOPs, STaTS offers an exciting alternative for efficient and robust time series classification. 

\subsection{Multivariate Classification}

We next evaluate STaTS on the multivariate classification task using the UEA-30 archive. Table~\ref{tab:uea_compression_comparison} presents results comparing STaTS-based summarization to both original TS2Vec and alternative compression strategies. As expected, TS2Vec (ori) achieves the highest overall accuracy (0.690) and the best average rank (3.39), operating on full-resolution inputs. However, \textit{TS2Vec (mean)} delivers competitive performance with an accuracy of 0.622 and a rank of 4.70, while operating on inputs that are on average \textit{20× shorter} (8.2 vs. 163.4 time steps). This substantial compression comes at only a modest cost in classification performance, highlighting the effectiveness of STaTS in preserving task-relevant features even in high-dimensional settings.

Among the compressed variants, TS2Vec (mean) outperforms both TS2Vec (GMM) and TS2Vec (uniform), which achieve lower accuracies of 0.589 and 0.488, and significantly worse ranks of 5.52 and 7.26, respectively. These results reinforce that simple fixed-interval segmentation (as in uniform) or clustering-based summarization (as in GMM) struggle to preserve discriminative temporal patterns in multivariate signals. The inferior performance of uniform segmentation is particularly pronounced, indicating that equal-length chunking fails to align with meaningful temporal structure. GMM-based summarization, while expressive, suffers from unstable representations, especially in short or low-variance windows.

\begin{table}[t]
\centering
\resizebox{0.475\textwidth}{!}{
\begin{tabular}{lccccc}
\toprule
\textbf{Dataset} & \textbf{Horizon} & \textbf{TS2Vec (ori)} & \textbf{TS2Vec (mean)} & \textbf{Informer} & \textbf{TCN} \\
\midrule
\multirow{2}{*}{ETTh1} 
  & 24   & 0.599 & 0.840 & \textbf{0.577} & 0.767 \\
  & 720  & \textbf{1.048} & 1.166 & 1.215 & 1.453 \\
\midrule
\multirow{2}{*}{ETTh2} 
  & 24   & \textbf{0.398} & 0.430 & 0.720 & 1.365 \\
  & 720  & 2.650 & \textbf{2.647} & 3.467 & 3.690 \\
\midrule
\multirow{2}{*}{ETTm1} 
  & 24   & \textbf{0.443} & 0.541 & 0.323 & 0.324 \\
  & 288  & 0.709 & \textbf{0.918} & 1.056 & 1.270 \\
\midrule
\multirow{2}{*}{Electricity} 
  & 24   & \textbf{0.287} & 0.312 & 0.312 & 0.297 \\
  & 720  & \textbf{0.375} & 0.873 & 0.969 & 0.447 \\
\bottomrule
\end{tabular}
}
\caption{
Comparison of normalized MSE for TS2Vec variants against Informer and TCN on three multivariate and one univariate (Electricity) forecasting datasets.
\textbf{TS2Vec (mean)} maintains strong performance at long horizons despite 15$\times$ compression and even outperforms Informer and TCN on most long-range forecasts.
}
\label{tab:forecasting_compression}
\end{table}

An analysis of the sequence length distribution in UEA-30 supports this observation: while the median sequence length is 292, the archive includes highly variable inputs, ranging from short sequences ($<$100 steps) to extremely long ones (e.g., \textit{EigenWorms} with 17{,}984 steps). Such diversity amplifies the limitations of static chunking strategies and highlights the value of STaTS, which adapts segmentation to local statistical changes within each sequence. These results show that \textit{STaTS-based summarization generalizes effectively to multivariate inputs}, offering a scalable and robust approach to reducing sequence length while maintaining accuracy across diverse datasets.

\subsection{Time Series Forecasting}

To evaluate the effect of summarization on time series forecasting, we follow the protocol outlined in TS2Vec~\cite{yue2022ts2vec}. Given a window of $L$ historical observations, a model is trained to predict the next $H$ values, where $H$ is the forecast horizon. Instead of directly training a regression model, TS2Vec treats forecasting as a downstream task built on top of its pre-trained encoder. A linear layer is trained atop the frozen representations to predict future values. We use this approach to assess how well representations learned from compressed sequences support forecasting accuracy. Experiments are conducted on four widely-used benchmark datasets: ETTh1, ETTh2, ETTm1, and Electricity (univariate), across a range of short- and long-term horizons ($H \in \{24, 288, 720\}$). We report normalized Mean Squared Error (nMSE), which normalizes the MSE by the variance of the ground truth, enabling comparisons across datasets and scales.

\begin{figure}[t]
    \centering
    \includegraphics[width=0.98\linewidth]{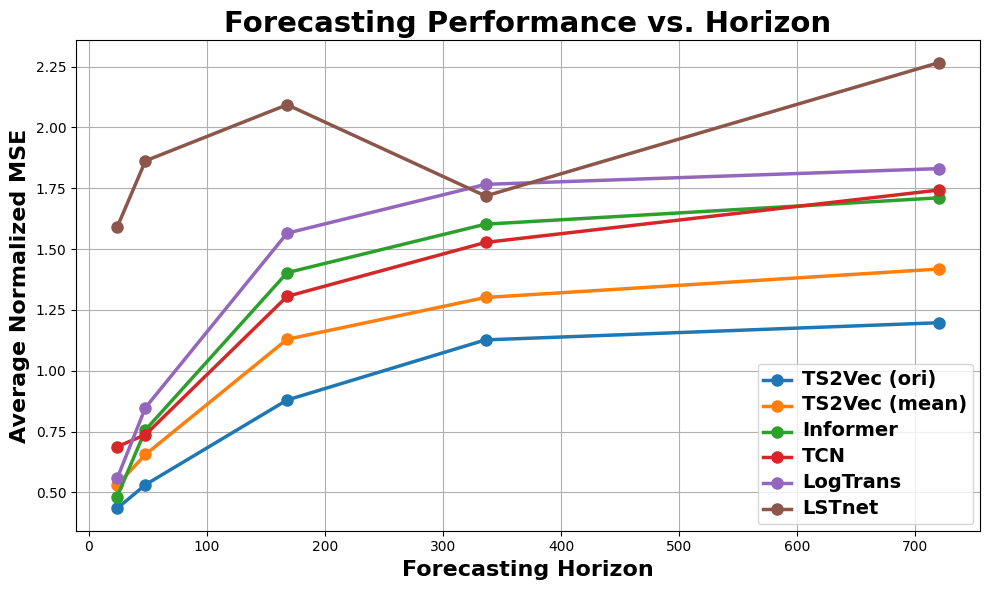}
    \caption{Average normalized MSE across increasing forecast horizons on four multivariate datasets. TS2Vec (mean), using STaTS-based summarization, shows strong long-term performance and outperforms Informer, TCN, and LogTrans beyond 300 steps, despite using inputs compressed over 15×. LSTnet degrades significantly at longer horizons, while TS2Vec (ori) remains the strongest short-range model.}
    \label{fig:forecasting_horizon}
    \Description[Forecasting error by horizon]{A line plot showing normalized mean squared error (nMSE) across increasing forecast horizons for different models. TS2Vec (mean) shows strong performance at long horizons.}
\end{figure}

Table~\ref{tab:forecasting_compression} shows that \textbf{TS2Vec (mean)} achieves strong forecasting performance despite operating on inputs that are over \textbf{15$\times$ shorter} than those used in the original TS2Vec model. While it slightly underperforms TS2Vec (ori) at short horizons (e.g., 24 steps), it consistently closes the gap or outperforms it at longer horizons, especially on ETTh2 and Electricity. Notably, at horizon 720, TS2Vec (mean) outperforms both Informer and TCN on ETTh2, and matches or surpasses TS2Vec (ori) on multiple datasets. This suggests that the STaTS-based summarization acts as an implicit smoothing mechanism, effectively capturing long-term trends while filtering out high-frequency noise—an advantage particularly valuable in long-horizon forecasting. On structured or low-variance datasets like ETTh2, TS2Vec (mean) demonstrates clear robustness and generalization, outperforming more complex baselines. These results highlight the potential of STaTS not just for compression, but also as signal denoising that enhances long-range temporal modeling.

\textbf{Scaling Forecasting Accuracy with Horizon.} Figure~\ref{fig:forecasting_horizon} highlights the impact of increasing forecast horizons on model accuracy. While TS2Vec (ori) achieves the best performance at short horizons, its compressed counterpart TS2Vec (mean) closes the gap rapidly and even surpasses several baselines, including Informer and TCN, at longer ranges ($>336$ steps). This suggests that STaTS summarization reduces input dimensionality and provides an inductive bias that enhances trend capture over extended periods. Unlike LogTrans~\cite{li2019enhancing} and LSTnet~\cite{lai2018modeling}, which deteriorate sharply at large horizons, TS2Vec (mean) maintains stable degradation, reinforcing its value for resource-constrained long-range forecasting tasks. The consistent gap between TS2Vec (mean) and TS2Vec (uniform) highlights the importance of structure-aware summarization. 

\begin{figure}[t]
  \centering
  \begin{tabular}{cc}
  \includegraphics[width=0.2\textwidth]{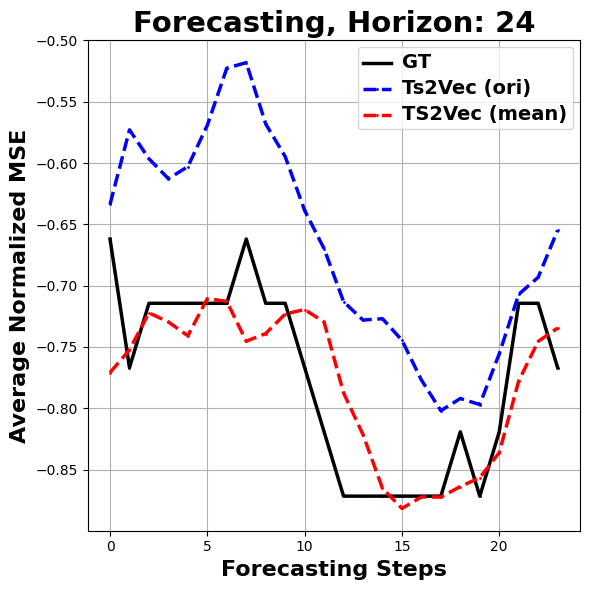} & 
  \includegraphics[width=0.2\textwidth]{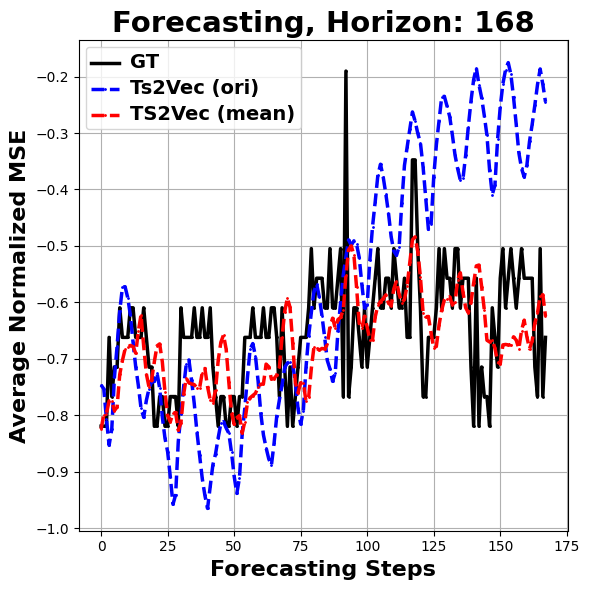} \\
  (a) & (b)\\
  \includegraphics[width=0.2\textwidth]{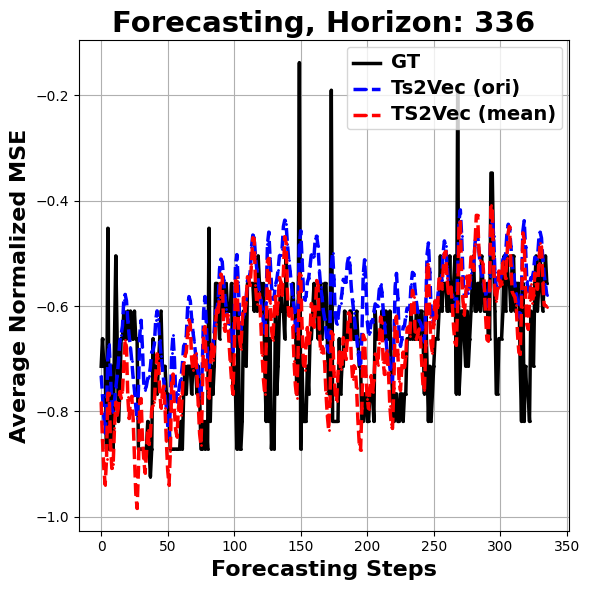} & 
  \includegraphics[width=0.2\textwidth]{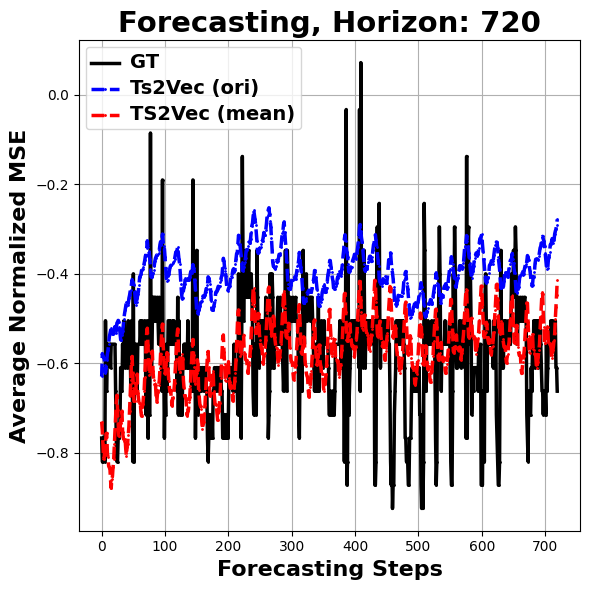}\\
  (c) & (d)\\
  \end{tabular}
  \caption{Qualitative forecasts at (a) short (H=24), (b) mid (H=168), (c) long (H=336), and (d) very long (H=720) horizons on the Electricity dataset. TS2Vec trained with STaTS (red) better tracks the ground truth (black) than the original TS2Vec (blue), particularly at longer horizons.}
  \label{fig:forecast_viz}
  \Description[Qualitative forecasting comparisons]{Four subplots showing predicted vs. ground truth time series on the Electricity dataset at different forecasting horizons (24, 168, 336, 720). TS2Vec (mean) tracks the signal more closely at longer horizons.}

\end{figure}

\subsection{Qualitative Analysis}
\textbf{Long-Horizon Forecasting.} To better understand how STaTS-based summarization affects forecast quality, we present qualitative comparisons across different prediction horizons on the Electricity dataset in Figure~\ref{fig:forecast_viz}. These examples span short (H=24), medium (H=168), long (H=336), and very long (H=720) horizons, providing insight into how different models behave as temporal uncertainty increases. 
At shorter horizons (H=24), both TS2Vec variants produce forecasts generally aligned with the ground truth, though STaTS (red, dashed) already shows a slightly smoother and more stable prediction path. As the horizon increases to H=168 and beyond, the advantage of STaTS becomes significantly more pronounced. While the original TS2Vec model (blue, dashed) begins to exhibit overconfident, oscillatory artifacts that diverge from the signal trend, STaTS maintains a trajectory that closely follows the ground truth, even capturing key transitions and plateaus. This behavior illustrates how structure-aware summarization helps the model generalize better under distributional uncertainty by filtering out noise and reducing spurious temporal variance. 
At the longest horizon (H=720), STaTS tracks the overall signal shape more accurately and remains more robust to sharp fluctuations that cause TS2Vec (ori) to drift substantially. These results support the earlier quantitative findings in Table~\ref{tab:forecasting_compression}, where STaTS outperformed or matched the baseline on long-range metrics. Importantly, this qualitative evidence highlights that summarization is not just a compression tool, but a means to induce temporal inductive bias, yielding more interpretable and stable behavior when extrapolating into the future.

\begin{figure}
    \centering
    \includegraphics[width=0.75\linewidth]{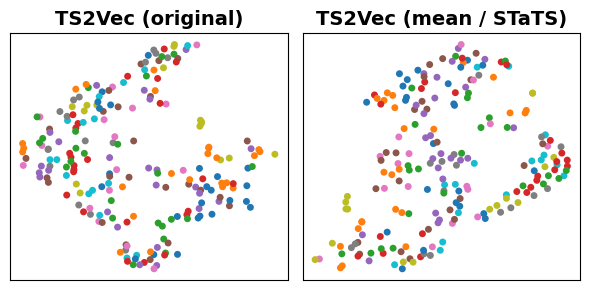}
    \caption{\textit{t-SNE visualization of learned representations on the GestureMidAirD3 dataset}. Despite operating on highly compressed inputs, TS2Vec (mean) with STaTS (right) preserves clear class separation and compactness comparable to the original TS2Vec (left), suggesting that summarization retains task-relevant structure while reducing redundancy.}
    \label{fig:tsne_gunpoint}
    \Description[t-SNE visualization of embeddings]{Side-by-side t-SNE plots showing representation quality on the GestureMidAirD3 dataset. TS2Vec (mean) achieves tight intra-class clustering comparable to full-resolution TS2Vec.}

\end{figure}
\textbf{Representation Quality.} To assess how summarization affects the quality of learned representations, we visualize the t-SNE projections of sample embeddings from the GestureMidAirD3 dataset for both TS2Vec (original) and TS2Vec (mean) in Figure~\ref{fig:tsne_gunpoint}. Each point corresponds to a time series sample, colored by its class label. Despite operating on inputs compressed over 30×, TS2Vec (mean) preserves the overall class separation structure and even achieves slightly tighter intra-class clusters than its full-resolution counterpart. This suggests that STaTS maintains discriminative information through compression and may act as an implicit denoising step, leading to more compact and separable representations.

\subsection{Discussion}

The results across classification and forecasting tasks demonstrate that STaTS-based summarization is a practical and robust solution for time series compression. A key finding is that STaTS enables models to operate on inputs that are 10--30$\times$ shorter than the original sequences while retaining 85--90\% of their predictive performance. This trade-off is especially compelling for scenarios where computational efficiency is critical—such as deployment on edge devices or training at scale across large datasets.

\textit{When should one use STaTS?} Our experiments show that STaTS excels under three conditions: (1) when input sequences are long and irregular, where uniform chunking fails to align with semantic boundaries; (2) when signals contain significant noise, where STaTS provides implicit denoising via statistical aggregation; and (3) in resource-constrained environments where full-resolution training is impractical. In both the UCR-128 archive and long-horizon forecasting benchmarks, STaTS consistently narrowed the gap with full-resolution models while offering dramatic input reduction. Conversely, when sequences are very short or highly regular, the benefits of adaptive segmentation diminish, and uniform chunking performs comparably at lower complexity.

\textit{How does STaTS compare to other compression strategies?} TS2Vec (mean) outperforms both TS2Vec (uniform) and TS2Vec (GMM) across all benchmarks. Uniform segmentation, though simple, is brittle: it assumes evenly distributed temporal structure and often misaligns with transitions, leading to degraded performance. GMM-based summarization, while more expressive, suffers from unstable clustering in short or noisy segments and performs poorly on fine-grained classification tasks. STaTS balances adaptivity and stability by leveraging local statistical divergence to guide segmentation without additional model parameters or supervision.

\textit{Assumptions and Limitations.} While STaTS is designed to be lightweight and model-agnostic, it assumes local statistical coherence within segments and may underperform when dynamics are abrupt or chaotic across all scales. Unlike end-to-end learned pooling or attention mechanisms, it does not use gradient-based feedback to adapt its compression. However, this design enables strong generalization and efficient deployment in resource-constrained settings.

% Overall, STaTS can be viewed as a lightweight inductive prior that infuses time series learning pipelines with structure-aware compression. Its plug-and-play compatibility with existing backbones like TS2Vec and demonstrated robustness across domains make it a promising approach for efficient time series representation learning. Future work could explore integrating STaTS into end-to-end training or combining it with attention mechanisms to improve expressivity under compression.

\section{Conclusion}\label{sec:conclusion}
This work presents STaTS, a general-purpose framework for Structure-Aware Temporal Summarization that enables time series data to be compressed into compact, information-preserving representations without compromising downstream performance. Unlike conventional approaches that rely on fixed-window segmentation or uniform subsampling, STaTS uses a principled, unsupervised strategy to detect change points based on the Bayesian Information Criterion (BIC), capturing transitions in local dynamics across multiple temporal resolutions. Each segment is then summarized using simple yet expressive statistics (e.g., means or generative tokens), producing a fixed-length sequence that preserves the structural essence of the original signal. 
We integrate STaTS with TS2Vec and demonstrate that this lightweight preprocessing step enables models to achieve comparable or superior performance to full-resolution baselines on univariate and multivariate classification and long-horizon forecasting while reducing input length by up to 30×. Beyond efficiency, STaTS also offers enhanced robustness to noise, outperforming both uniform and clustering-based summarization baselines in noisy settings. Our results generalize across three major benchmarks, UCR, UEA, and ETT, highlighting that adapting model input to the data’s intrinsic temporal structure is a powerful inductive bias for efficient, resilient learning. 
These findings open several promising directions for future work. First, extending STaTS to online or streaming settings would support real-time summarization and edge deployment. Second, integrating STaTS with multimodal or hierarchically structured data (e.g., sensor networks or video) could enable more holistic event abstraction. Finally, incorporating STaTS into end-to-end adaptive learning systems—especially under distribution shift or concept drift—offers a path toward resilient time series modeling in non-stationary, real-world environments. Ultimately, we aim to extend STaTS as a compression tool and a scalable, structure-aware front-end for modern time series understanding.
% \clearpage
% \section{GenAI Usage Disclosure}
% LLMs such as ChatGPT and Gemini were used for proofreading and correcting grammatical issues found during writing and debugging during code development. The listed contributors authored all technical content and analysis.

\textbf{Acknowledgements.} This work was supported by the US NSF grants IIS 2348689 and IIS 2348690, and the USDA award no. 2023-69014-39716.

\bibliographystyle{ACM-Reference-Format}
\bibliography{egbib}

\end{document}